\documentclass[11pt,a4paper]{article}
\usepackage{emnlp2021}
\pdfoutput=1
\usepackage{times}
\usepackage{latexsym}

\usepackage{colortbl}
\usepackage{booktabs}
\usepackage{sidecap}
\usepackage{amsmath}
\usepackage{amssymb}
\usepackage{enumitem}
\usepackage{color, colortbl}
\usepackage{multirow}
\usepackage{xcolor}
\usepackage{sidecap}
\usepackage{graphicx}
\usepackage{anyfontsize}
\usepackage[breaklinks]{hyperref}
\usepackage{breakurl}

\usepackage{microtype}

\newcommand{\Fref}[1]{Figure~\ref{#1}}

\newcommand{\Tref}[1]{Table~\ref{#1}}

\newcommand\pz{\hphantom{0}}
\newcommand\parnospace[1]{\noindent\textbf{#1}}

\defcitealias{oeis}{OEIS}

\title{Learning Mathematical Properties of Integers}

\author{Maria Ryskina\thanks{\ \ Research performed at DiDi Labs.} \\
  Language Technologies Institute \\
  Carnegie Mellon University \\
  \texttt{\fontsize{10}{10}\selectfont mryskina@cs.cmu.edu} \\\And
  Kevin Knight \\
  DiDi Labs \\
  \texttt{\fontsize{10}{10}\selectfont kevinknight@didiglobal.com} \\}

\date{}

\begin{document}
\maketitle

\begin{abstract}
Embedding words in high-dimensional vector spaces has proven valuable in many natural language applications. In this work, we investigate whether similarly-trained embeddings of integers can capture concepts that are useful for mathematical applications. We probe the integer embeddings for mathematical knowledge, apply them to a set of numerical reasoning tasks, and show that by learning the representations from mathematical sequence data, we can substantially improve over number embeddings learned from English text corpora.
\end{abstract}

\section{Introduction}

Word vector representations learned by neural models have been shown to capture linguistic knowledge that can be useful for downstream NLP tasks \cite{belinkov2019analysis}.
In this work, we look at whether similarly-trained integer embeddings can represent useful mathematical knowledge.
As a simple example, a mathematician may look at a few values of the function $2^n-1$:

\vspace{0.1in}

\noindent

\begin{small}
\begin{tabular}{|c|l|l|r|r|r|r|r|r|} \hline
$n$ & 1 & 2 & 3 & 4 & 5 & 6 & 7 & ... \\ \hline
$2^n-1$ & 1 & 3 & 7 & 15 & 31 & 63 & 127 & ... \\ \hline
\end{tabular}
\end{small}

\vspace{0.1in}

\noindent
and notice patterns such as \emph{`if $n$ is even, then $2^n-1$ is divisible by $3$'}. This type of reasoning is frequent in the early stages of mathematical work.
In order for software to assist people in identifying such patterns, it needs to have an internal representation of the basic properties of integers.  Just as learned word representations can capture attributes like the word's part of speech or syntactic dependency label \cite{kohn2015whats, belinkov-etal-2017-neural}, we would like to develop integer embeddings that capture primality, divisibility by 3, etc.

In this paper, we learn integer embeddings from mathematical resources and probe them for mathematical knowledge. Unlike most natural-language features, many number-theoretic properties are highly regular and often may be easily determined from the integer value itself (e.g.\ converting an integer to its binary representation instantly reveals whether it is a power of 2). However, our focus is not on building classifiers to predict these properties, but rather on exploring whether this information can be learned from the integer co-occurrence statistics in mathematical data and how it is distributed across the vector's dimensions.

Besides that, we are also interested in whether the integer representations learned from context can capture meaningful mathematical properties without us having to define them explicitly. It is not clear how one would probe for such unspecified features, but we can instead estimate the usefulness of the learned embeddings for downstream numerical reasoning tasks. In this work, we apply several types of integer embeddings to a set of tasks built around mathematical regularities, and show that embeddings learned from integer sequences yield better performance than the ones trained on text.

\begin{figure}[t]
\small
\texttt{[A000040]}: 
2, 3, 5, 7, 11, 13, 17, 19, 23, 29, ... \hfill (primes)\\
\texttt{[A000045]}: 
0, 1, 1, 2, 3, 5, 8, 13, 21, ... \hfill (Fibonacci)\\
\texttt{[A000108]}:
1, 1, 2, 5, 14, 42, 132, 429, ... \hfill (Catalan)\\
\texttt{[A005132]}:
0, 1, 3, 6, 2, 7, 13, 20, 12, 21, ... \hfill (Recamán)

\caption{Example sequences and their IDs in the Online Encyclopedia of Integer Sequences (OEIS).}
\label{fig:oeis-examples}
\end{figure}

\begin{SCtable*}[0.7][t]
    \centering \small
    \begin{tabular}{llrr}
    \toprule
Method & Corpus & Tokens &  
Types \\ 
\midrule
Ours & OEIS & 14M & 133,004 \\ 
\midrule
GloVe--840B--300D & Common Crawl & 840B & 64,729 \\ 
SkipGram--BoW--5 & Wikipedia & $\sim$2.5B & 2,272 \\ 
FastText--Wiki & Wikipedia, UMBC, \texttt{statmt.org} & 16B & 12,037 \\ 
    \bottomrule
\end{tabular}
\caption{Training corpus details for the integer embeddings used in this work. OEIS types here only include integers that occur 3 or more times; all others are replaced by UNK.}
\label{tab:embeddings}
\end{SCtable*}

\section{Data: OEIS}

Our source of mathematical knowledge is the Online Encyclopedia of Integer Sequences \citepalias{oeis}, a well-known database of number sequences representing properties that are of interest to mathematicians~\cite{sloane2003online}. Several samples from OEIS are shown in
\Fref{fig:oeis-examples}. While some OEIS sequences may be recognized by non-mathematicians (e.g.\ the Fibonacci sequence \texttt{\small [A000045]}), others are based on complex mathematical regularities and would be extremely difficult for a layperson to interpret (e.g.\ the greatest possible number of diagonals of a polyhedron with $n$ faces \texttt{\small [A279015]}).

The sequential structure of OEIS allows us to use it for training NLP models developed for textual data, such as recurrent neural network (RNN) language models and co-occurrence-based word embeddings. Of the 336K sequences in OEIS, we allocate 90\% for training and divide the rest into development (used for hyperparameter tuning) and test splits; split statistics are listed in \Tref{tab:oeis-summary}. Of the 1.5M integer types in the vocabulary, 83\% appear in training only once. The sequences are represented by their $n$ first elements ($n=43$ on average), so the coverage of larger integers in OEIS is sparser, and they make up most of the out-of-vocabulary items in the development and test sets.\footnote{For the OEIS data download details, see Appendix~\ref{app:data}.}

\section{Integer Embeddings}
In our experiments, we compare the integer embeddings learned from OEIS with the representations of integer tokens in the vocabulary of word embedding models trained on English text. \Tref{tab:embeddings} details the training corpus statistics for all embeddings.

\begin{table}[t]
    \centering
    \resizebox{\columnwidth}{!}{
    \begin{tabular}{lrrr}
    \toprule
 & Train & Dev & Test \\
    \midrule
    Sequences & 302,281  & 16,793 & 16,793 \\
    Integer tokens & 12,979,924 & 719,808 & 719,237\\
    Mean sequence length & 43 & 43 & 43 \\
    Integer types & 1,531,064 & 128,488 & 127,976 \\
    Singleton types & 1,268,771 & -- & -- \\
    Token OOV rate & 0\% & 10.0\% & 9.8\% \\
    \bottomrule
 \end{tabular}
 }
    \caption{Summary of the OEIS data splits used in this work. The development split is used for model selection and the test split is used in the sequence completion experiments. Type statistics are reported without minimum count filtering. Singleton types refer to the integers that occur only once in the training set.}
    \label{tab:oeis-summary}
\end{table} 

\begin{table*}[t]
\centering \small
\begin{tabular}{lcccccccc}
\toprule
\multirow{2}{*}{Embeddings} & \multicolumn{2}{c}{Evenness} & \multicolumn{2}{c}{Divisibility by 3} & \multicolumn{2}{c}{Divisibility by 4} & \multicolumn{2}{c}{Primality} \\
\cmidrule(lr){2-3} \cmidrule(lr){4-5} \cmidrule(lr){6-7} \cmidrule(lr){8-9}
& Single & All & Single & All & Single & All & Single & All \\ 
\midrule
Random baseline & 0.50 & 0.50 & 0.67 & 0.67 & 0.75 & 0.75 & 0.87 & 0.87 \\
\midrule
GloVe--840B--300D & 0.51 & 0.76 & 0.67 & 0.47 &0.75 & 0.71 & 0.87 & 0.84\\
SkipGram--BoW--5  & 0.50 & 0.52 & 0.67 & 0.67 & 0.75 & 0.75 & 0.87 & 0.87\\
FastText--Wiki & 0.51 & 0.61 & 0.67 & 0.66 & 0.75 & 0.76 & 0.87 & 0.62\\
\midrule
OEIS--LSTM & 0.69 & 0.93 & 0.67 & 0.71 &0.76 & 0.81  & 0.86 & 0.95\\
OEIS--LSA & {\bf 0.80} & 0.62 & 0.67 & 0.67 & 0.75 & 0.75 & 0.87 & 0.87\\
OEIS--FastText (with subword units) & 0.78 & {\bf 1.00} & {\bf 0.69} & 0.94 & {\bf 0.80} & {\bf 1.00} & 0.82 & {\bf 1.00}\\
OEIS--FastText (no subword units)  & 0.59 & {\bf 1.00} & 0.68 & {\bf 0.98} & 0.77 & 0.98 & 0.86 & {\bf 1.00}\\
\midrule
Concatenate(OEIS--FastText, FastText--Wiki) & 0.78 & {\bf 1.00} & {\bf 0.69} & 0.94 & {\bf 0.80} & {\bf 1.00} & 0.82 & {\bf 1.00} \\
\bottomrule
\end{tabular}
\caption{Accuracies of the logistic regression probing classifiers for three binary properties. The classifiers are trained on integers 1--1000, and results are reported on 1001--2000. ``All'' uses the entire integer embedding as input, while ``Single'' uses only the most relevant component of the embedding, chosen by train (1--1000) accuracy. Highest accuracy in each experiment is shown in \textbf{bold}, if it is exceeds the random choice baseline.}
\label{probe1}
\end{table*}

\begin{table*}
\centering \small
\begin{tabular}{lcccc}
\toprule
\multirow{2}{*}{Embeddings} & \multicolumn{2}{c}{Value ($R^2$)} & \multicolumn{2}{c}{Magnitude ($R^2$)} \\
\cmidrule(lr){2-3}\cmidrule(lr){4-5}
& Single & All & Single & All  \\ 
\midrule
    GloVe--840B--300D  & 0.82 & {\bf 1.00} & 0.65 & {\bf 0.99}\\
    SkipGram--BoW--5  & 0.79 & 0.99 & 0.70 & 0.97\\
    FastText--Wiki  & {\bf 0.90} & 0.99 & {\bf 0.76} & 0.98 \\
    \midrule
    OEIS--LSTM  & 0.49 & 0.86 & 0.31 & 0.78 \\
    OEIS--LSA  & 0.08 & 0.55 & 0.13 & 0.53 \\
    OEIS--FastText (with subword units) & 0.38 & 0.99 & 0.33 & 0.96 \\
    OEIS--FastText (no subword units) & 0.49 & 0.99 & 0.39 & 0.95 \\
    \midrule
    Concatenate(OEIS--FastText, FastText--Wiki) & {\bf 0.90} & {\bf 1.00} & {\bf 0.76} & 0.98 \\
    \bottomrule
\end{tabular}
\caption{Performance of the linear regression models trained to predict an integer's value and order of magnitude (number of digits). We fit the regression models to integers 1--2000 and evaluate the coefficient of determination $R^2$ of the fit on the same set (between 0 and 1, higher is better). Best results in each experiment shown in \textbf{bold}.} 
\label{probe2}
\end{table*}

\subsection{Embeddings Learned from OEIS}
\label{oeis-embeddings}

Motivated by the distributional semantics paradigm, we learn multi-dimensional embedding vectors from the contexts in which integers occur in OEIS. We use three training methods, well-known in NLP: 

\smallskip
\parnospace{LSTM embeddings:} we train a two-layer LSTM language model on the 300K OEIS training sequences and extract the 100-dimensional vectors from the weight matrix of its embedding layer.\footnote{For implementation details, see Appendix~\ref{app:lstm}.}

\vspace{0.1cm}
\parnospace{LSA embeddings:} Latent Semantic Analysis \cite{hofmann1999probabilistic} performs  dimensionality reduction on a ``document-term'' matrix using truncated singular value decomposition. The rows of the matrix represent the integer sequences (as available in OEIS), the columns stand for the integer types, and the values in the cells are equal to the number of occurrences of each integer in each sequence. The resulting matrix is sparse, and its low rank yields vectors with a maximum of 65 dimensions. This method takes OEIS co-occurrences into account, but not the ordering within the sequences.

\vspace{0.1cm}
\parnospace{FastText:} we learn 100-dimensional 
FastText embeddings~\cite{bojanowski-etal-2017-enriching}, both with and without
the additional subword-level information.


\subsection{Pre-trained Embeddings}
\label{pre-trained-embeddings}

Prior work has successfully employed word representations that are pre-trained on large text corpora for NLP applications. Naturally, these text corpora include numerical data as well, and some properties of a number can be inferred from the textual context (e.g., if an integer follows the phrase \emph{the year}, it most likely has four digits). While the training data for these embeddings is not designed to encode integer properties, it may contain billions of tokens, compared to only $\sim$13M in OEIS.
It has been shown that word embeddings retain some knowledge of the integer properties~\cite{naik-etal-2019-exploring}, so we exploit these resources as a baseline.

We use three sets of pre-trained embeddings: GloVe~\cite{pennington-etal-2014-glove}, SkipGram bag-of-words~\cite{levy-goldberg-2014-dependency}, and FastText~\cite{bojanowski-etal-2017-enriching}. For all three models, we select the versions that performed best in the numerical knowledge tests of~\citet{naik-etal-2019-exploring}.\footnote{Specific model versions and links to download them are listed in Appendix~\ref{app:pretrained}.}   

\newcite{thawani-etal-2021-representing} provide a comprehensive description of prior work on number representations learned from text. Prior work concentrates on basic numeracy (counting, paraphrasing, relative magnitude, word problems), while we focus on integer properties relevant to mathematical fields such as number theory.

\section{What's in an Integer Embedding?}

For the learned vectors to be useful for mathematical applications, they need to contain the information about the important integer properties, and this information can be distributed over one or several neurons. For example, in the LSTM embeddings trained on OEIS, we quickly find an ``evenness neuron'':  the 156th element of the embedding vector $v$ is generally positive for even numbers and negative for odd numbers. This holds for integers from 1 to 50 with only a few exceptions, e.g.: 

{\small
\begin{align*}
v_{156}(\textcolor{blue}{1}) &= \textcolor{red}{\hphantom{-}0.15} &v_{156}(\textcolor{red}{6}) &= \textcolor{red}{\hphantom{-}0.38}\\
v_{156}(\textcolor{red}{2}) &= \textcolor{red}{ \hphantom{-}0.29} &v_{156}(\textcolor{blue}{7}) &= \textcolor{blue}{-0.31}\\
v_{156}(\textcolor{blue}{3}) &= \textcolor{blue}{-0.04} & v_{156}(\textcolor{red}{8}) &= \textcolor{red}{\hphantom{-}0.39} \\
v_{156}(\textcolor{red}{4}) &= \textcolor{red}{\hphantom{-}0.26} & v_{156}(\textcolor{blue}{9}) &= \textcolor{blue}{-0.02}\\
v_{156}(\textcolor{blue}{5}) &= \textcolor{blue}{ -0.08} & v_{156}(\textcolor{red}{10}) &= \textcolor{red}{\hphantom{-}0.43}
\end{align*}
}

\vspace{-0.6cm}
\paragraph{Probing classifiers.} 
We use probing classifiers to test whether the embeddings learn mathematical properties. To avoid spurious correlations, we keep our classifiers very simple.
We start by probing for three binary properties: divisibility by 2, 3, and 4, and primality. We train a logistic regression classifier on integers from 1 to 1000 and evaluate its predictions on integers from 1001 to 2000. 

\begin{table*}[!t]
    \small
    \centering
    \begin{tabular}{p{3.5cm}@{}c@{\hspace{1cm}}l}
        \toprule
        Prompt & Answer & Explanation \\
        \midrule
         0, 2, 4, 6, & \pz 8 & $a_{n+1} = a_{n} + 2$ (arithmetic progression)\\
         0, 1, 1, 2, 3, 5, & \pz 8 & $a_{n+2} = a_{n} + a_{n+1}$ (Fibonacci numbers) \\
         65536, 256, 16, & \pz 4 & $a_{n+1} = \sqrt{a_n}$\\
         6, 28, 12, 14, 24, 7, & 48 & $a_{2n+1} = 2\cdot a_{2n-1}, a_{2n+2} = a_{2n} / 2$ (alternating geometric progressions)\\
         32, 35, 39, 44, & 50 & $a_{n+1} = a_{n} + n + 2$ (increments forming an arithmetic progression)\\
        \bottomrule
    \end{tabular}
    \caption{Example numerical sequence completion problems used in our experiments, along with the rationales behind the gold answers. All questions shown here are collected from the test preparation website \href{https://www.nibcode.com/en/psychometric-training/test-of-numerical-sequence}{\texttt{\small Nibcode}}. In our experimental setup, the language model encodes the prompt and needs to predict the most likely continuation.}
    \label{tab:sequence_completion_data}
\end{table*}
\definecolor{light-gray}{gray}{0.9}
\newcolumntype{g}{>{\columncolor{light-gray}}c}
\begin{table*}[ht]
    \small
    \centering
    \begin{tabular}{r@{ : }r@{\hspace{0.2cm}::\hspace{0.2cm}}r@{ : }r@{\hspace{0.7cm}}ggg@{\hspace{0.5cm}}l}
    \toprule
     \emph{a} & \emph{b} & \emph{c} & ? & \multicolumn{3}{l}{Incorrect answer options}& Explanation\\
     \midrule
     5 & 36 & 6 & 49 & 48 & 50 & 56 & $x : (x+1)^2$\\
     42 & 56 & 72 & 90 & 81 & 92 & 100 & $x(x+1) : (x+1)(x+2)$ \\ 
     48 & 122 & 168 & 290 & 215	& 225 & 292 & $(x^2-1) : [(x+4)^2+1]$\\
     210 & 380 & 182 & 342 & 156 & 240 & 272 & $(x^2-x) : [(x+5)^2-(x+5)]$\\
     11529 & 7235 & 152943 & 213549 & 62034& 163044 & 203448 & The sum of the digits in each pair is the same\\
    \bottomrule
    \end{tabular}
    \caption{Sample mathematical analogy problems used in our experiments, along with the rationales behind the gold answers; the questions shown here are collected from the test preparation website \href{https://learnfrenzy.com/reasoning/verbal-reasoning/verbal-analogy/number-analogy/}{\texttt{\small LearnFrenzy}}. All analogy problems are of the form \emph{a:b :: c:?}, and the task is to choose one of the 3--5 provided options that would most appropriately fill in the blank. The column labeled ``?'' shows the expected correct answer, and the numbers shaded in grey are the remaining answer options. In our experiments with integer embeddings, we use vector arithmetic to find the projected vector of the answer and then select its nearest neighbor among the presented options.
    }
    \label{tab:analogy_data}
\end{table*}

Table~\ref{probe1} shows the prediction accuracies for classifiers trained on each embedding type, using either the entire vector or the single most predictive dimension. 
Generally speaking, the classifiers trained on  LSA vectors and pre-trained word embeddings perform on par with the random baseline, providing no evidence of containing the relevant information. However, the OEIS-trained LSTM and FastText vectors can be reliably used to predict mathematical properties of integers, strongly outperforming the baseline in all cases.  

Subword information (in our case, digits) is also useful for inferring mathematical properties: e.g., divisibility by 4 can be determined from the last two digits of a number. The subword-based version of FastText averages vectors for all digit 3-grams to 6-grams, plus the vector for the whole integer. We get further improvements by training classifiers on a concatenation of OEIS--FastText and FastText--Wiki vectors, both trained with subword units.

We also probe for the number's order of magnitude (number of digits) and for the value of the integer itself using a linear regression (Table~\ref{probe2}): we fit it on integers 1--2000 and evaluate the coefficient of determination $R^2$ of the fit. In this case, vectors pre-trained on large text corpora perform well. 

\paragraph{Completing integer sequences.} The task of predicting the next integer in a given sequence has long been used in aptitude testing as a measure of inductive reasoning skills.\footnote{We note the non-uniqueness of solutions is a persistent criticism of these tests~\cite{korossy1998solvability}.}
We collect 57 sequence completion problems from online test preparation websites offering practice questions. \Tref{tab:sequence_completion_data} shows five sample sequence completion problems from our dataset, annotated with answer explanations.\footnote{For the full list of sequence completion problems and their sources, see \url{https://github.com/ryskina/integer-embedding-tests}}

Most sequences test the subject's ability to recognize common patterns like arithmetic progression ({\small 0, 2, 4, 6, ? $\rightarrow$ 8}) or well-known sequences like Fibonacci numbers ({\small 0, 1, 1, 2, 3, 5, ? $\rightarrow$ 8}), but some sequences feature additional distractor patterns such as multiple distinct sequences alternating or the increments themselves forming a sequence following another pattern ({\small 32, 35, 39, 44, ? $\rightarrow$ 50}).

\Tref{tab:seq-completion} compares the performance of the LSTM model trained on OEIS with two baselines:

\textbf{Baseline 1: GPT-2.} We predict the next token using the OpenAI GPT-2~\cite{radford2019language} language model.\footnote{We use the AllenNLP prediction demo: \url{https://demo.allennlp.org/next-token-lm}} GPT-2  can memorize sequences very well, but it is not specifically trained for mathematical generalization. GPT-2 achieves slightly higher accuracy (Precision@1) than our LSTM model but substantially lower Precision@5.  

\textbf{Baseline 2: Search.} We match the prompt directly against the OEIS database and return the most frequent continuation.  This is a strong baseline: 56\% of the test set questions occur somewhere in OEIS followed by the gold answer at least once. 

While human aptitude questions focus on a specific range of patterns that are relatively easy for people to identify, we also want to test the predictive abilities on the OEIS test set sequences which showcase more sophisticated phenomena. 
The LSTM embeddings yield higher accuracy in that case (lower half of \Tref{tab:seq-completion}).
The search baseline is not as effective for this task because the prompts are now much longer (42 tokens on average compared to 5). Limiting the search to only the last 5 tokens of the prompt improves accuracy, but it still does not outperform the OEIS--LSTM.

\begin{table}
\small
    \centering
    \begin{tabular}{llrr}
    \toprule
    Test set & Method & P@1 & P@5 \\ 
    \midrule
    Aptitude tests 
    & OEIS--LSTM & 0.05 & 0.37 \\ 
    & GPT-2 & 0.07 & 0.19 \\
    & OEIS search & {\bf 0.53} & {\bf 0.56}\\
    \midrule
    OEIS test set
    & OEIS--LSTM & {\bf 0.14} & {\bf 0.26} \\ 
    & OEIS search (full) & 0.02 & 0.02\\
    & OEIS search (last 5) & 0.12 & 0.17 \\
    \bottomrule
    \end{tabular}
    \caption{Performance on the sequence completion task. Precision@$k$ measures how often the reference answer is among the top $k$ predicted continuations. For the OEIS search baseline, we sort the possible next tokens by frequency of occurring after the given sequence prefix (using either the full prefix or its last five elements).}
    \label{tab:seq-completion}
\end{table}

\paragraph{Mathematical analogies.} A traditional benchmark for evaluating word representations is word analogy, i.e. answering questions of the form ``\emph{a} is to \emph{b} as \emph{c} is to ...'' \cite{mikolov2013efficient}. We perform a similar mathematical analogy test, collecting 79 questions from numerical aptitude practice tests.
All questions are multiple-choice (3 to 5 answer options), to mitigate non-uniqueness of the solution. \Tref{tab:analogy_data} shows five sample analogy problems from our dataset, annotated with the intended analogy explanations.\footnote{Full list of mathematical analogy problems and their sources can be found at \url{https://github.com/ryskina/integer-embedding-tests}}

Following \citet{mikolov-etal-2013-linguistic}, we use vector arithmetic to solve analogies: out of the given options, for the task \emph{a:b :: c:?} we choose the number whose embedding has the highest cosine similarity to the vector $v(c)-v(a)+v(b)$.
\Tref{tab:analogy} shows that only the FastText embeddings learned from OEIS outperform the random choice baseline on this task. We believe this is due to the linear regularity encoded in the embeddings by design (lacking in LSA and LSTM), and the training data that highlights mathematical properties useful for the task (as compared to FastText--Wiki).

\begin{table}
  \small
    \centering
    \begin{tabular}{lr}
    \toprule
    Method & Accuracy \\ 
    \midrule
    Random choice baseline & 0.28 \\
    OEIS--LSTM & 0.25\\ 
    OEIS--LSA & 0.27\\
    OEIS--FastText & {\bf 0.34} \\
    FastText--Wiki & 0.18\\
    \bottomrule
    \end{tabular}
    \caption{Peformance on the multiple-choice numerical analogy questions. The answer to \emph{a:b :: c:?} is chosen by highest cosine similarity to $v(c)-v(a)+v(b)$. OEIS--FastText is trained with subword information.}
    \label{tab:analogy}
\end{table}

\paragraph{Expanding integer seed sets.} In mathematical exploration, we often have a small set $S$ of integers that exhibit some behavior and aim to find other integers that behave similarly. We test the models' ability to expand a given set by finding the centroid of the embeddings of $S$'s members and sorting candidates by cosine distance to the centroid.

Some qualitative examples are shown in Table~\ref{seed-sets}.  Given the seed set {\bf 5~13~29}, top results given by the OEIS-trained FastText include small prime numbers, while for {\bf 73~97~83} we mostly get prime numbers of similar magnitude.  The seed {\bf 729~1024~243} returns other powers of~2 and 3. By contrast, embeddings trained on non-OEIS texts mainly just return additional numbers in the same range.

\begin{table}
\small
\centering
\begin{tabular}{lll}
\toprule
Seed set & Vectors & Candidate expansions \\
\midrule
5 13 29 &
OEIS--FT & {\bf 19, 7, 17, 3,} 9, {\bf 23} \\
& FT--Wiki & 26, 12, 28, 14, 27, 15  \\
\midrule
73 97 83 & 
OEIS--FT & {\bf 79, 71, 67, 89,} 77, {\bf 103} \\
& FT--Wiki & 82, 81, 78, 84, 76, {\bf 79} \\
\midrule
729 1024 243 & 
OEIS--FT & {\bf 2187, 81, 256, 64, 27, 512} \\
& FT--Wiki & 768, {\bf 256}, 640, 840, 384, 216 \\
\bottomrule
\end{tabular}
\caption{Given a small seed set of integers, we predict other candidate members of the set, ranked by the cosine similarity of their vectors to the centroid of the seed set. Methods include OEIS--FastText and FastText--Wiki, both trained with subword information.}
\label{seed-sets}
\end{table}

\section{Conclusion}

We introduce integer embeddings trained on the Online Encyclopedia of Integer Sequences (OEIS) and probe them for mathematical knowledge along with the integer representations found in the vocabulary of pre-trained English word embeddings. We find the OEIS embeddings promising for mathematical applications, as they are able to capture some meaningful mathematical regularities.

\section*{Acknowledgements}

We thank the DiDi Labs NLP Group members for helpful discussion, and the anonymous reviewers for their valuable feedback.

\bibliographystyle{acl_natbib}
\bibliography{custom,anthology}

\appendix
\section{Data sources}
\label{app:data}

\parnospace{OEIS} sequence dump was downloaded from the encyclopedia website: \url{https://oeis.org/stripped.gz} (accessed on 2020-07-22). 
\medskip

\noindent\textbf{Sequence completion} problems were collected from five test preparation websites: \href{https://www.nibcode.com/en/psychometric-training/test-of-numerical-sequence}{\texttt{\small Nibcode}}, \href{https://www.syvum.com/cgi/online/serve.cgi/iq/ar_series3.html?question_hide}{\texttt{\small Syvum}},
\href{https://www.12minprep.com/knowledge-hub/number-series-practice/#sample-questions-horizontal}{\texttt{\small 12MinPrep}}, \href{https://iqtestprep.com/number-sequence-test/}{\texttt{\small IQTestPrep}}, \href{https://www.hitbullseye.com/Number-Series-Tricks.php}{\texttt{\small Hitbullseye}}.
\medskip

\noindent\textbf{Mathematical analogy} problems were collected from four test practice websites: \href{https://learnfrenzy.com/reasoning/verbal-reasoning/verbal-analogy/number-analogy/}{\texttt{\small LearnFrenzy}}, \href{https://www.examsbook.com/number-analogy-reasoning-questions}{\texttt{\small Examsbook}}, \href{https://www.toppr.com/guides/reasoning-ability/analogy/number-analogy/}{\texttt{\small toppr}}, \href{http://www.allindiaexams.in/reasoning/verbal-reasoning-questions-answers/number-analogies}{\texttt{\small AllIndiaExams}}.

\section{Pre-trained embeddings}
\label{app:pretrained}

Based on the empirical analysis of numeracy in word embeddings conducted by \citet{naik-etal-2019-exploring}, we choose three embedding models pre-trained on English text corpora:
\begin{itemize}
\item \textbf{GloVe--840B--300D:} 300-dimensional GloVe vectors trained on 840B tokens from the Common Crawl corpus. Downloaded from \url{https://nlp.stanford.edu/projects/glove}
\item \textbf{SkipGram--BoW--5:} 300-dimensional SkipGram bag-of-words embeddings with window size 5. Downloaded from \url{https://levyomer.wordpress.com/2014/04/25/dependency-based-word-embeddings}
\item \textbf{FastText--Wiki:} 300-dimensional FastText vectors with subword information, trained on the 16B tokens from Wikipedia 2017, UMBC webbase corpus and \texttt{statmt.org} news dataset. Downloaded from \url{https://fasttext.cc/docs/en/english-vectors.html}
\end{itemize}

\section{LSTM implementation details}
\label{app:lstm}
We implement our two-layer LSTM language model using the PyTorch toolkit. Our model uses truncated BPTT \cite{williams1990efficient} and is trained for 40 epochs. Hyperparameters: hidden size 200, dynamically annealed learning rate starting at 20, gradients clipped at 0.25.

\end{document}